\documentclass{article} % For LaTeX2e
\usepackage{nips12submit_e,times}
\usepackage{url,epsfig}

\title{Efficient Estimation of Word Representations in Vector Space}

\author{
Tomas Mikolov \\
Google Inc.,
Mountain View, CA \\
\texttt{tmikolov@google.com} \\
\And
Kai Chen \\
Google Inc.,
Mountain View, CA \\
\texttt{kaichen@google.com} \\
\AND
Greg Corrado \\
Google Inc.,
Mountain View, CA \\
\texttt{gcorrado@google.com} \\
\And
Jeffrey Dean \\
Google Inc.,
Mountain View, CA \\
\texttt{jeff@google.com} \\
}

\nipsfinalcopy % Uncomment for camera-ready version

\begin{document}

\renewcommand{\arraystretch}{1.3}

\maketitle

\begin{abstract}
We propose two novel model architectures for computing continuous vector representations of words from very large data sets.
The quality of these representations is measured in a word similarity task, and the results are compared to the previously best performing techniques based on different types of neural networks.
We observe large improvements in accuracy at much lower computational cost, i.e. it takes less than a day to learn high quality word vectors
from a 1.6 billion words data set. Furthermore, we show that these
vectors provide state-of-the-art performance on our test set for measuring syntactic and semantic word similarities.
\end{abstract}

\section{Introduction}

Many current NLP systems and techniques treat words as atomic units - there is no notion of similarity between words, as these are represented as indices in a vocabulary.
This choice has several good reasons - simplicity, robustness and the observation that simple models trained on huge amounts of data
outperform complex systems trained on less data. An example is the popular N-gram model used for statistical language modeling - today, it is possible to
train N-grams on virtually all available data (trillions of words~\cite{stupid-backoff}).

However, the simple techniques are at their limits in many tasks. For example, the amount of relevant in-domain data for automatic speech recognition is limited - the performance is usually dominated by
the size of high quality transcribed speech data (often just millions of words). In machine translation, the existing corpora for many languages contain only a few billions of words or less.
Thus, there are situations where simple scaling up of the basic techniques will not result in any significant progress, and we have to focus on more advanced techniques.

With progress of machine learning techniques in recent years, it has become possible to train more complex models on much larger data set, and they typically outperform the simple models.
Probably the most successful concept is to use distributed representations of words~\cite{Hinton}.
For example, neural network based language models significantly outperform N-gram models~\cite{Bengio, Schwenk, MikolovIS2011}.

%In this paper, we focus on the increasingly popular technique of representing words in a continuous vector space, which may overcome data sparsity problems that arise when words are treated as atomic units.
%A good motivation for considering continuous vector representations of words might be the statistical language modeling, because similar words can share parameters, which leads to more robust estimation of the parameters during training.
%Compared to N-gram models that treat each word as an atomic unit, this approach provides significantly better results~\cite{Bengio, Schwenk, MikolovIS2011}.

\subsection{Goals of the Paper}

The main goal of this paper is to introduce techniques that can be used for learning high-quality word vectors from huge data sets with billions of words, and with millions of words in the vocabulary.
As far as we know, none of the previously proposed architectures has been successfully trained on more than a few hundred of millions of words, with a modest dimensionality of the word vectors between 50 - 100.

%We propose two new architectures for estimating continuous word vectors that closely follow observation that shallow architecture is sufficient for estimating the word vectors~\cite{dip}.

We use recently proposed techniques for measuring the quality of the resulting vector representations, with the expectation that not only
will similar words tend to be close to each other, but that words can have {\bf multiple degrees of similarity}~\cite{NAACL1}. This has been observed earlier in the context of inflectional
languages - for example, nouns can have multiple word endings, and if we search for similar words in a subspace of the original vector space, it is possible to find words that have similar endings~\cite{dip, Mikolov}.

Somewhat surprisingly, it was found that similarity of word representations goes beyond simple syntactic regularities. Using a word offset technique where simple algebraic operations are performed on the
word vectors, it was shown for example that {\it vector("King") - vector("Man") + vector("Woman")} results in a vector that is closest to the vector representation of the word {\it Queen}~\cite{NAACL1}.

In this paper, we try to maximize accuracy of these vector operations by
developing new model architectures that preserve the linear regularities among words. We design a new comprehensive test set for measuring both
syntactic and semantic regularities\footnote{The test set is available at \url{www.fit.vutbr.cz/~imikolov/rnnlm/word-test.v1.txt}},
and show that many such regularities can be learned
with high accuracy. Moreover, we discuss how training time and accuracy depends on the dimensionality of the word vectors and on the amount of the training data.

\subsection{Previous Work}

Representation of words as continuous vectors has a long history~\cite{Hinton, BPTT, Elman}. A very popular model architecture for estimating neural network language model (NNLM) was proposed in~\cite{Bengio},
where a feedforward neural network with a linear projection layer and a non-linear hidden layer was used to learn jointly the word vector representation and a statistical language model. This work has been
followed by many others.

Another interesting architecture of NNLM was presented in~\cite{dip, Mikolov}, where the word vectors are first learned using neural network with a single hidden layer. The
word vectors are then used to train the NNLM. Thus, the word vectors are learned even without constructing the full NNLM.
In this work, we directly extend this architecture, and focus just on the first step where the word vectors are learned using a simple model.

It was later shown that the word vectors can be used to significantly improve and simplify many NLP applications~\cite{Collobert1, Collobert2, Turian}. Estimation of the word vectors itself was
performed using different model architectures and trained on various corpora~\cite{Collobert1, Turian, Mnih, thesis, Huang}, and some of the resulting word vectors were made
available for future research and comparison\footnote{\url{http://ronan.collobert.com/senna/} \\ \url{http://metaoptimize.com/projects/wordreprs/} \\ \url{http://www.fit.vutbr.cz/~imikolov/rnnlm/} \\ \url{http://ai.stanford.edu/~ehhuang/} }.
However, as far as we know, these architectures were significantly more computationally expensive for training
than the one proposed in~\cite{dip}, with the exception of certain version of log-bilinear model where diagonal weight matrices are used~\cite{Mnih}.

\section{Model Architectures}

Many different types of models were proposed for estimating continuous representations of words, including the well-known Latent Semantic Analysis (LSA) and Latent Dirichlet Allocation (LDA). In this paper, we focus
on distributed representations of words learned by neural networks, as it was previously shown that they perform significantly better than LSA for preserving linear regularities among words~\cite{NAACL1, NAACL2}; LDA moreover becomes computationally
very expensive on large data sets.

Similar to~\cite{ASRU}, to compare different model architectures we define first the computational complexity of a model as the number of parameters that need to be accessed to fully train the model. Next, we will
try to maximize the accuracy, while minimizing the computational complexity.

For all the following models, the training complexity is proportional to
\begin{equation}
O = E \times T \times Q,
\end{equation}
where $E$ is number of the training epochs, $T$ is the number of the words in the training set and $Q$ is defined further for each model architecture. Common choice is $E = 3 - 50$ and $T$ up to one billion.
All models are trained using stochastic gradient descent and backpropagation~\cite{BPTT}.

\subsection{Feedforward Neural Net Language Model (NNLM)}

%- normal NNLM description, with projection layer used to derive the word vectors

The probabilistic feedforward neural network language model has been proposed in~\cite{Bengio}. It consists of input, projection, hidden and output layers. At the input layer, $N$ previous words are encoded
using 1-of-$V$ coding, where $V$ is size of the vocabulary. The input layer is then projected to a projection layer $P$ that has dimensionality $N \times D$, using a shared projection matrix. As only $N$ inputs are
active at any given time, composition of the projection layer is a relatively cheap operation.

The NNLM architecture becomes complex for computation between the projection and the hidden layer, as values in the projection layer are dense.
For a common choice of $N = 10$, the size of the projection layer ($P$) might be 500 to 2000, while the hidden layer size $H$ is typically 500 to 1000 units. Moreover, the hidden layer is used to compute probability
distribution over all the words in the vocabulary, resulting in an output layer with dimensionality $V$. Thus, the computational complexity per each training example is
\begin{equation}
Q = N \times D + N \times D \times H + H \times V,
\end{equation}
where the dominating term is $H \times V$. However, several practical solutions were proposed for avoiding it; either using hierarchical versions of the softmax~\cite{Morin, Mnih, ASRU}, or avoiding
normalized models completely by using models that are not normalized during training~\cite{Collobert1, Huang}. With binary tree representations of the vocabulary, the number of output units that need to be evaluated
can go down to around $log_{2}(V)$. Thus, most of the complexity is caused by the term $N \times D \times H$.

In our models, we use hierarchical softmax where the vocabulary is represented as a Huffman binary tree.
This follows previous observations that the frequency of words works well for obtaining classes in neural net language models~\cite{MikolovICASSP2011}.
Huffman trees assign short binary codes to frequent words, and this further reduces the number of output units that need to be evaluated: while balanced binary tree would require $log_{2}(V)$
outputs to be evaluated, the Huffman tree based hierarchical softmax requires only about $log_{2}(Unigram\_perplexity(V))$. For example when the vocabulary size is one million words, this results in about two times speedup in evaluation.
While this is not crucial speedup for neural network LMs as the computational bottleneck is in the $N \times D \times H$ term, we will later propose architectures that do not have hidden layers and thus depend
heavily on the efficiency of the softmax normalization.

\subsection{Recurrent Neural Net Language Model (RNNLM)}

Recurrent neural network based language model has been proposed to overcome certain limitations of the feedforward NNLM, such as the need to specify the context length (the order of the model $N$), and because
theoretically RNNs can efficiently represent more complex patterns than the shallow neural networks~\cite{MikolovIS, BengioAI}. The RNN model does not have a projection layer; only input, hidden and output layer.
What is special for this type of model is the recurrent matrix that connects hidden layer to itself, using time-delayed connections. This allows the recurrent model to form some kind of short term memory, as information
from the past can be represented by the hidden layer state that gets updated based on the current input and the state of the hidden layer in the previous time step.

The complexity per training example of the RNN model is
\begin{equation}
Q = H \times H + H \times V,
\end{equation}
where the word representations $D$ have the same dimensionality as the hidden layer $H$. Again, the term $H \times V$ can be efficiently reduced to $H \times log_{2}(V)$ by using hierarchical softmax. Most of the complexity
then comes from $H \times H$.

%\section{Hierarchical Softmax}

%- cite [Morin Bengio], [Mnih Hinton], then propose Huffman tree to get the binary codes for words using frequencies to maximize speed
% Huffman tree - from log2(V) to log2(unigram PPL) - for 1 million vocab, this goes from 20 to maybe 11?

\subsection{Parallel Training of Neural Networks}

To train models on huge data sets, we have implemented
several models on top of a large-scale distributed framework called
DistBelief~\cite{DistBelief},
including the feedforward NNLM and the new models proposed in this
paper. The framework allows us to run multiple replicas of the same
model in parallel, and each
replica synchronizes its gradient updates through a centralized
server that keeps all the parameters. For this parallel training, we
use mini-batch asynchronous
gradient descent with an adaptive
learning rate procedure called Adagrad~\cite{AdaGrad}. Under this framework, it is common to use one
hundred or more model replicas, each using many CPU cores at different machines in a data center.

\section{New Log-linear Models}

%- minimization of the training complexity

%- Model Training - SGD with decreasing learning rate

In this section, we propose two new model architectures for learning distributed representations of words that try to minimize computational complexity.
The main observation from the previous section was that most of the complexity is caused by the non-linear hidden layer in the model. While this is what makes neural networks so attractive, we decided to explore simpler models that might
not be able to represent the data as precisely as neural networks, but can possibly be trained on much more data efficiently. % Note this is not in contradiction with our goal to use more complex models as it was set in the
%introduction: the following models use continuous distributed representations of words, in contrast to local representations used by models such as N-grams.

The new architectures directly follow those proposed in our earlier work~\cite{dip, Mikolov}, where it was found that neural network language model can be successfully trained in two steps: first, continuous word vectors are learned using simple
model, and then the N-gram NNLM is trained on top of these distributed representations of words. While there has been later substantial amount of work that focuses on learning word vectors, we consider the approach proposed
in~\cite{dip} to be the simplest one. Note that related models have been proposed also much earlier~\cite{BPTT, Elman}.

\subsection{Continuous Bag-of-Words Model}

\begin{figure}[tb]
\centering
\centerline{\epsfig{figure=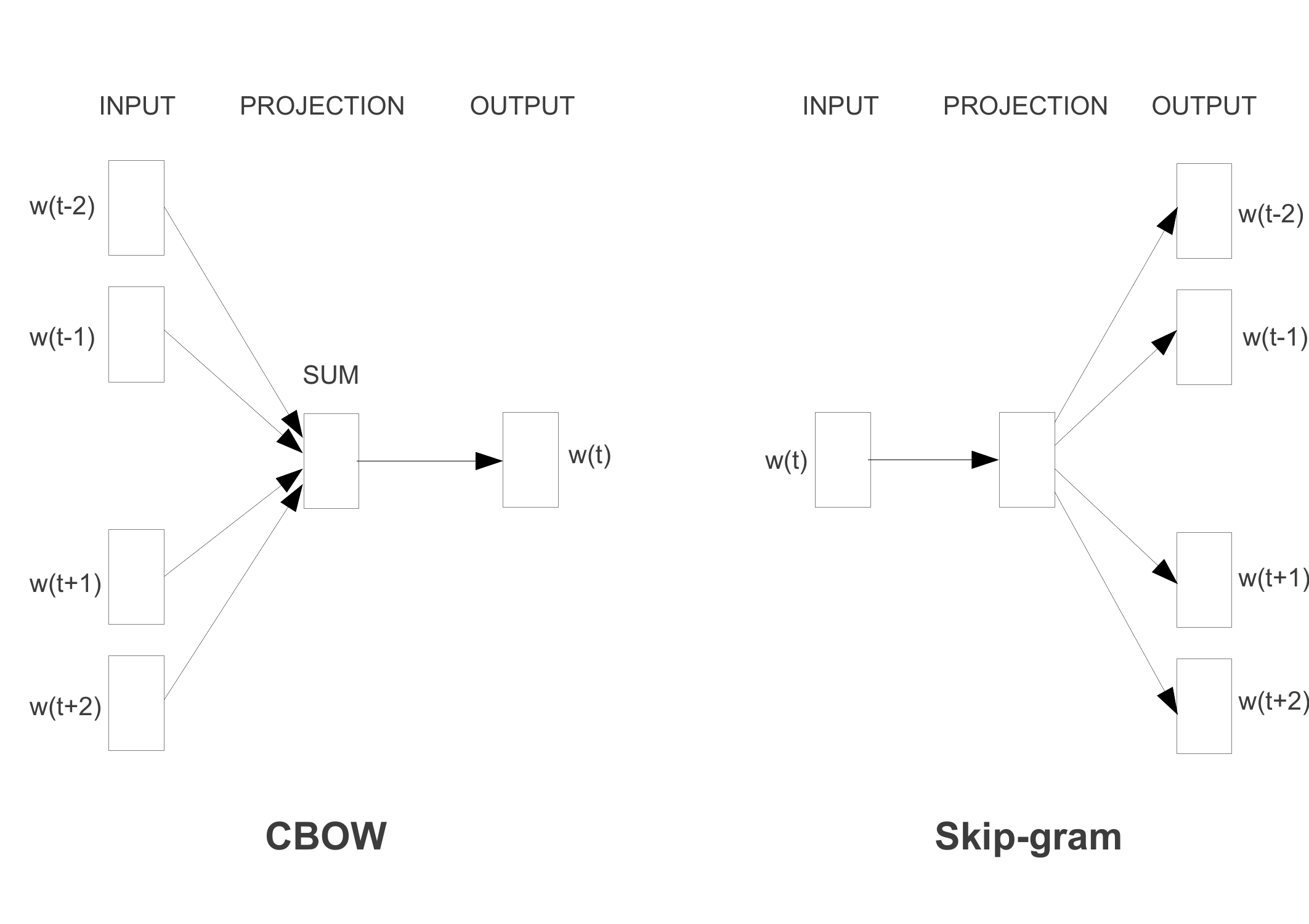,width=130mm}}
\caption{New model architectures. The CBOW architecture predicts the current word based on the context, and the Skip-gram predicts surrounding words given the current word.}
\label{fig-models}
\end{figure}

The first proposed architecture is similar to the feedforward NNLM, where the non-linear hidden layer is removed and the projection layer is shared for all words (not just the projection matrix); thus, all words get projected
into the same position (their vectors are averaged). We call this architecture a bag-of-words model as the order of words in the history does not influence the projection.
Furthermore, we also use words from the future; we have obtained the best performance on the task introduced in the next section by building a log-linear classifier with four future and four history words at the input,
where the training criterion is to correctly classify the current (middle) word. Training complexity is then
\begin{equation}
Q = N \times D + D \times log_{2}(V).
\label{eq1}
\end{equation}
We denote this model further as CBOW, as unlike standard bag-of-words model, it uses continuous distributed representation of the context. The model architecture is shown at Figure~\ref{fig-models}. Note that the weight
matrix between the input and the projection layer is shared for all word positions in the same way as in the NNLM.

\subsection{Continuous Skip-gram Model}

The second architecture is similar to CBOW, but instead of predicting the current word based on the context, it tries to maximize classification of a word based on another word in the same sentence. More precisely, we use each current
word as an input to a log-linear classifier with continuous projection layer, and predict words within a certain range before and after the current word. We found that increasing the range improves quality of the resulting word vectors,
but it also increases the computational complexity. Since the more distant words are usually less related to the current word than those close to it, we give less weight to the distant words by sampling less from
those words in our training examples.% To address this fact, we can weight importance of more distant words to be lower. We used a more effcient solution, where the range of words is uniformly random, thus more distant words are less often presented as examples.

The training complexity of this architecture is proportional to
\begin{equation}
Q = C \times (D + D \times log_{2}(V)),
\end{equation}
where $C$ is the maximum distance of the words. Thus, if we choose $C=5$, for each training word we will select randomly a number $R$ in range $<1 ; C>$, and then use $R$ words from history and $R$ words
from the future of the current word as correct labels. This will require us to do $R \times 2$ word classifications, with the current word as input, and each of the $R + R$ words as output.
In the following experiments, we use $C=10$.

\section{Results}

To compare the quality of different versions of word vectors, previous papers typically use a table showing example words and their most similar words, and understand them intuitively.
Although it is easy to show that word {\it France} is similar to {\it Italy} and perhaps some other countries, it is much more challenging when subjecting those vectors in a more complex similarity task, as follows.
We follow previous observation that there can be many different types of similarities between words, for example, word {\it big} is similar to {\it bigger} in the same sense that
{\it small} is similar to {\it smaller}. Example of another type of relationship can be word pairs {\it big - biggest} and {\it small - smallest}~\cite{NAACL1}. We further denote two pairs of words with the same relationship
as a question, as we can ask: "What is the word that is similar to {\it small} in the same sense as {\it biggest} is similar to {\it big}?"
%We can take two pairs of words with the same relationship
%and ask the following question: "What is the word that is similar to {\it small} in the same sense as {\it biggest} is similar to {\it big}?".

Somewhat surprisingly, these questions can be answered
by performing simple algebraic operations with the vector representation of words. To find a word that is similar to {\it small} in the same sense as {\it biggest} is similar to {\it big}, we can simply compute
vector $X = vector("biggest") - vector("big") + vector("small")$. Then, we search in the vector space for the word closest to $X$ measured by cosine distance, and use it as the answer to the question (we discard the input
question words during this search).
When the word vectors are well trained, it is possible to find the correct answer (word $smallest$) using this method.% The better we can answer this question, the beter the quality of the vectors is.

Finally, we found that when we train high dimensional word vectors on a large amount of data, the resulting vectors can be used to answer very subtle semantic relationships between words, such as a city and the
country it belongs to, e.g. France is to Paris as Germany is to Berlin. Word vectors with such semantic relationships could be used to improve many existing NLP applications, such as machine translation, information retrieval
and question answering systems, and may enable other future applications yet to be invented. %We intend to publish a set of high quality word vectors trained on large amounts of data later.

\subsection{Task Description}

%- description of the task: word relationship test, syntactic + semantic questions

To measure quality of the word vectors, we define a comprehensive test set that contains five types of semantic questions, and nine types of syntactic questions. Two examples from each category are shown in Table~\ref{tab1}. Overall,
there are 8869 semantic and 10675 syntactic questions. The questions in each category were created in two steps: first, a list of similar word pairs was created manually. Then, a large list of questions is formed by connecting
two word pairs. %This large number of questions was created semi-automatically, by connecting word pairs from manually written sets of word pairs.
For example, we made a list of 68 large American cities and the states they belong to, and formed about 2.5K questions by picking two word pairs at random. We have included in our test set only single token words,
thus multi-word entities are not present (such as {\it New York}). %We plan to publish the test set soon, so that other researchers can use it in their experiments.

We evaluate the overall accuracy for all question types, and for each question type separately (semantic, syntactic). Question is assumed to be correctly answered only if the closest
word to the vector computed using the above method is exactly the same as the correct word in the question; synonyms are thus counted as mistakes. This also means that reaching 100\% accuracy is likely to
be impossible, as the current models do not have any input information about word morphology.
However, we believe that usefulness of the word vectors for certain applications should be positively correlated with this accuracy metric. Further progress can be achieved by incorporating
information about structure of words, especially for the syntactic questions.

\begin{table} [tb]
\caption{ {\it Examples of five types of semantic and nine types of syntactic questions in the Semantic-Syntactic Word Relationship test set.}}
\vspace{2mm}
\centerline{
\begin{tabular}{|l||c|c||c|c|}
\hline
Type of relationship           & \multicolumn{2}{c||}{Word Pair 1}  &     \multicolumn{2}{c|}{Word Pair 2}\\
\hline
Common capital city &    Athens       &      Greece     &    Oslo     &    Norway      \\
All capital cities  &    Astana       &      Kazakhstan &    Harare   &    Zimbabwe    \\
Currency            &    Angola       &      kwanza     &    Iran     &    rial        \\
City-in-state       &    Chicago      &      Illinois   &    Stockton &    California   \\
Man-Woman           &    brother      &      sister     &    grandson &    granddaughter    \\
\hline
Adjective to adverb &   apparent      &    apparently   &    rapid    &    rapidly       \\
Opposite            &   possibly      &    impossibly   &    ethical  &    unethical     \\
Comparative         &   great         &    greater      &    tough    &    tougher       \\
Superlative         &   easy          &    easiest      &    lucky    &    luckiest      \\
Present Participle  & think           & thinking        &   read      &   reading        \\
Nationality adjective &  Switzerland  &    Swiss        &   Cambodia  &   Cambodian      \\
Past tense          &   walking       &   walked        &   swimming  &   swam           \\
Plural nouns        &   mouse         &    mice         &   dollar    &   dollars        \\
Plural verbs        &   work          &         works   &    speak    &   speaks         \\
\hline
\end{tabular}}
\label{tab1}
\end{table}

%- show results when the relationship vector is not used and we search just for the closest word

\subsection{Maximization of Accuracy}

We have used a Google News corpus for training the word vectors. This corpus contains about 6B tokens. We have restricted the vocabulary size to 1 million most frequent words. Clearly, we are facing
time constrained optimization problem, as it can be expected that both using more data and higher dimensional word vectors will improve the accuracy. To estimate the best choice of model architecture for obtaining
as good as possible results quickly, we have first evaluated models trained on subsets of the training data, with vocabulary restricted to the most frequent 30k words. The results using the CBOW
architecture with different choice of word vector dimensionality and increasing amount of the training data are shown in Table~\ref{Acc1}.

It can be seen that after some point, adding more dimensions or adding more training data provides diminishing improvements. So, we have to increase both vector dimensionality and the amount of the training data together. While
this observation might seem trivial, it must be noted that it is currently popular to train word vectors on relatively large amounts of data, but with insufficient size (such as 50 - 100).
% From Table~\ref{Acc1}, it is clear that using less data and more dimensionality helps, if vectors have too low dimensionality.
Given Equation~\ref{eq1}, increasing amount of training data twice results in about the same increase of computational complexity as increasing vector size twice.

%- time constrained optimization problem

\begin{table} [tb]
\caption{ {\it Accuracy on subset of the Semantic-Syntactic Word Relationship test set, using word vectors from the CBOW architecture with limited vocabulary. Only questions containing words from the most frequent 30k words are used.}}
\vspace{2mm}
\centerline{
\begin{tabular}{|c||c|c|c|c|c|c|}
\hline
Dimensionality / Training words         &  24M  &  49M  &  98M  &  196M  & 391M & 783M  \\
\hline
50                                      &  13.4 &  15.7 &  18.6 &  19.1  & 22.5 & 23.2 \\
100                                     &  19.4 &  23.1 &  27.8 &  28.7  & 33.4 & 32.2 \\
300                                     &  23.2 &  29.2 &  35.3 &  38.6  & 43.7 & 45.9 \\
600                                     &  24.0 &  30.1 &  36.5 &  40.8  & 46.6 & 50.4 \\  %783M / 600 with +-6 epochs gives 50.4, but didnt recompute this with 3 epochs - should be the same
\hline
\end{tabular}}
\label{Acc1}
\end{table}

For the experiments reported in Tables~\ref{Acc1} and~\ref{Acc2}, we used three training epochs with stochastic gradient descent and backpropagation. We chose starting learning rate 0.025 and decreased it linearly, so that it approaches zero
at the end of the last training epoch.

\subsection{Comparison of Model Architectures}

First we compare different model architectures for deriving the word vectors using the same training data and using the same dimensionality of 640 of the word vectors.
In the further experiments, we use full set of questions in the new Semantic-Syntactic Word Relationship test set, i.e. unrestricted to the 30k vocabulary.
We also include results on a test set introduced in~\cite{NAACL1} that focuses on syntactic
similarity between words\footnote{We thank Geoff Zweig for providing us the test set.}.

The training data consists of several LDC corpora and is described in detail in~\cite{ASRU} (320M words, 82K vocabulary).
We used these data to provide a comparison to a previously trained recurrent neural network language model
that took about 8 weeks to train on a single CPU. We trained a feedforward NNLM with the same number of 640 hidden units using the DistBelief parallel training~\cite{DistBelief}, using a history of 8 previous words (thus, the NNLM has
more parameters than the RNNLM, as the projection layer has size $640 \times 8$).

In Table~\ref{Acc0}, it can be seen that the word vectors from the RNN (as used in~\cite{NAACL1}) perform well mostly on the syntactic questions. The NNLM vectors perform significantly better than the RNN - this is not surprising, as the
word vectors in the RNNLM are directly connected to a non-linear hidden layer. The CBOW architecture
works better than the NNLM on the syntactic tasks, and about the same on the semantic one. Finally, the Skip-gram architecture works slightly worse on the syntactic task than the CBOW model (but still better than the NNLM), and much better
on the semantic part of the test than all the other models.

\begin{table} [tb]
\caption {{\it Comparison of architectures using models trained on the same data, with 640-dimensional word vectors. The accuracies are reported on our Semantic-Syntactic Word Relationship test set, and on the syntactic relationship test set of~\cite{NAACL1} }}
\vspace{2mm}
\centerline{
\begin{tabular}{|c||c|c||c|}
\hline
Model                      &    \multicolumn{2}{c||}{Semantic-Syntactic Word Relationship test set} &  MSR Word Relatedness  \\
\cline{2-3}
Architecture               &    Semantic Accuracy [\%]      &  Syntactic Accuracy [\%]                 &    Test Set~\cite{NAACL1} \\
\hline
RNNLM                      &  9  &       36    &       35  \\
NNLM                       & 23  &       53    &       47  \\
CBOW                       & 24  &       64    &       61  \\
Skip-gram                  & 55  &       59    &       56  \\
\hline
\end{tabular}}
\label{Acc0}
\end{table}

Next, we evaluated our models trained using one CPU only and compared the results against publicly available word vectors.
The comparison is given in Table~\ref{Acc2}. The CBOW model was trained on subset of the Google News data in about a day, while training time for the Skip-gram model was about three days.

\begin{table} [tb]
\caption{ {\it Comparison of publicly available word vectors on the Semantic-Syntactic Word Relationship test set, and word vectors from our models. Full vocabularies are used.}}
\vspace{2mm}
\centerline{
\begin{tabular}{|l||c|c||c|c|c|}
\hline
Model    &   Vector          & Training        &    \multicolumn{3}{c|}{Accuracy [\%]}  \\
         &   Dimensionality  &      words      &    \multicolumn{3}{c|}{}   \\
\hline
         &                             &       & Semantic      &  Syntactic    &  Total  \\
\hline
Collobert-Weston NNLM       &   50     & 660M  &   9.3 &  12.3 &  11.0 \\
Turian NNLM                 &   50     &  37M  &   1.4 &   2.6 &   2.1 \\
Turian NNLM                 &  200     &  37M  &   1.4 &   2.2 &   1.8 \\
Mnih NNLM                   &   50     &  37M  &   1.8 &   9.1 &   5.8 \\
Mnih NNLM                   &  100     &  37M  &   3.3 &  13.2 &   8.8 \\
Mikolov RNNLM               &   80     & 320M  &   4.9 &  18.4 &  12.7 \\
Mikolov RNNLM               &  640     & 320M  &   8.6 &  36.5 &  24.6 \\
Huang NNLM                  &   50     & 990M  &  13.3 &  11.6 &  12.3 \\
\hline
Our NNLM                    &   20     &   6B  &  12.9 &  26.4 &  20.3 \\
Our NNLM                    &   50     &   6B  &  27.9 &  55.8 &  43.2 \\
Our NNLM                    &  100     &   6B  &  34.2 & {\bf64.5} &  50.8 \\
CBOW                        &  300     & 783M  &  15.5 &  53.1 &  36.1 \\
Skip-gram                   &  300     & 783M  & {\bf 50.0} &  55.9 &  {\bf 53.3} \\
%These are with 30k vocab:
%Our NNLM                    &   20     &   6B  &  15.4 &  34.7 &  27.8 \\
%Our NNLM                    &   50     &   6B  &  28.2 &  68.3 &  50.8 \\
%Our NNLM                    &  100     &   6B  &  35.1 &  71.4 &  58.8 \\
%CBOW loglinear model        &  300     & 783M  &  20.7 &  59.7 &  45.9 \\
%Skip-gram loglinear model *2/3  &  300     & 783M  &  49.8 &  56.7 &  54.3 \\
\hline
\end{tabular}}
\label{Acc2}
\end{table}

For experiments reported further, we used just one training epoch (again, we decrease the learning rate linearly so that it approaches zero at the end of training).
Training a model on twice as much data using one epoch gives comparable or better results than iterating over the same data for three epochs,
as is shown in Table~\ref{Acc3}, and provides additional small speedup.

\begin{table} [tb]
\caption{ {\it Comparison of models trained for three epochs on the same data and models trained for one epoch. Accuracy is reported on the full Semantic-Syntactic data set.}}
\vspace{2mm}
\centerline{
\begin{tabular}{|l||c|c||c|c|c||c|}
\hline
Model    &   Vector          & Training        &    \multicolumn{3}{c||}{Accuracy [\%]}  & Training time \\
         &   Dimensionality  &      words      &    \multicolumn{3}{c||}{}               &    [days]     \\
\hline
         &                             &       & Semantic      &  Syntactic    &  Total &               \\
\hline
%Our NNLM                 &  100     &   6B  &  34.2 &  64.5 &  50.8             &    14 $\times$ 100    \\ % times 100? maybe more, 100 machines = 1000 CPUs?
3 epoch CBOW             &  300     & 783M  &  15.5 &  53.1 &  36.1             &     1     \\
3 epoch Skip-gram        &  300     & 783M  &  50.0 &  55.9 &  53.3             &     3    \\
\hline
1 epoch CBOW             &  300     & 783M  &  13.8 &  49.9 & 33.6                  &   0.3  \\
1 epoch CBOW             &  300     & 1.6B  &  16.1 &  52.6 & 36.1                  &   0.6  \\
1 epoch CBOW             &  600     & 783M  &  15.4 &  53.3 & 36.2                  &   0.7  \\
\hline
1 epoch Skip-gram        &  300     & 783M  &  45.6 &  52.2 & 49.2                  &   1   \\
1 epoch Skip-gram        &  300     & 1.6B  &  52.2 &  55.1 & 53.8                  &   2   \\
1 epoch Skip-gram        &  600     & 783M  &  56.7 &  54.5 & 55.5                  &   2.5 \\
\hline
\end{tabular}}
\label{Acc3}
\end{table}

\subsection{Large Scale Parallel Training of Models}

% *** TODO: add new best results - accuracy 67% or so - in a Table

As mentioned earlier, we have implemented various models in a
distributed framework called DistBelief. Below we report the
results of several models trained on the Google News 6B data set, with
mini-batch asynchronous gradient descent and the adaptive
learning rate procedure called Adagrad~\cite{AdaGrad}. We used 50 to 100 model
replicas during the training. % The NNLM is trained using 8 history words
%(9-gram) and three hidden layers of 768 neurons in each layer. The
%CBOW model is trained using 4 prefix and 4 postfix words. The
%SkipGram model is trained using a maximum skip of 10 words.
The number
of CPU cores is an estimate since the data center machines
are shared with other production tasks, and the usage can fluctuate
quite a bit. Note that due to the overhead of the distributed
framework, the CPU usage of the CBOW model and the Skip-gram model are
much closer to each other than their single-machine
implementations. The result are reported in Table~\ref{Acc4}.

% Raw Details:
%
% NNLM: batch-size 100, model replica: 50; param-server shards: 5;
%AdaGrad; 8 history words; three NN layers; hidden nodes: 768;
%ACTIVATION_RECTIFIED_LINEAR; Vocab: 982563; hinge-loss w/
%regularization 0.005; vector max_norm: 1.0.
% CBOW:  batch-size 100, model replica: 100; param-server shards: 10;
%AdaGrad; 4 prefix words + 4 postfix words; hinge-loss w/o
%regularization; no bouding of vectors.
% SKIP: batch-size 100, model replica: 100; param-server shards: 10;
%AdaGrad; max_skips: 10; hinge-loss w/o regularization; no bouding of
%vectors.
%
% CBOW: CPU: 4*10 + 100 = 140 cores. (100 replica)
% SKIP: CPU: 2.5*10 + 100= 125 cores (100 replica)
% NNLM: CPU: 1.5*5 + 3.5*50 = 180 cores. (50 replica)

\begin{table} [tb]
\caption{ {\it Comparison of models trained using the DistBelief
distributed framework.
Note that training of NNLM with 1000-dimensional vectors would take
too long to complete.}}
\vspace{2mm}
\centerline{
\begin{tabular}{|l||c|c||c|c|c||c|}
\hline
Model    &   Vector          & Training        & \multicolumn{3}{c||}{Accuracy [\%]}  & Training time \\
         &   Dimensionality  &      words      & \multicolumn{3}{c||}{}               &    [days x CPU cores]     \\
\hline
         &                  &       & Semantic      &   Syntactic    &  Total &               \\
\hline
NNLM             &  100     &   6B  &  34.2 &  64.5 &  50.8 &    14 x 180    \\
CBOW             & 1000     &   6B  &  57.3 &  68.9 &  63.7 &     2 x 140    \\
Skip-gram        & 1000     &   6B  &  66.1 &  65.1 &  65.6 &     2.5 x 125  \\
\hline
\end{tabular}}
\label{Acc4}
\end{table}

\subsection{Microsoft Research Sentence Completion Challenge}

The Microsoft Sentence Completion Challenge has been recently introduced as a task for advancing language modeling and other NLP techniques~\cite{MSRSCC}. This task consists of 1040 sentences, where
one word is missing in each sentence and the goal is to select word that is the most coherent with the rest of the sentence, given a list of five reasonable choices. Performance of several techniques has been already
reported on this set, including N-gram models, LSA-based model~\cite{MSRSCC}, log-bilinear model~\cite{Mnih2012} and a combination of recurrent neural networks that currently holds the state of the art
performance of 55.4\% accuracy on this benchmark~\cite{thesis}.

We have explored the performance of Skip-gram architecture on this task. First, we train the 640-dimensional model on 50M words provided in~\cite{MSRSCC}. Then, we compute score of each sentence in the test set by
using the unknown word at the input, and predict all surrounding words in a sentence. The final sentence score is then the sum of these individual predictions. Using the sentence scores, we choose the most likely
sentence.

A short summary of some previous results together with the new results is presented in Table~\ref{SCCTab}. While the Skip-gram model itself does not perform on this task better than LSA similarity,
the scores from this model are complementary to scores obtained with RNNLMs, and a weighted combination leads to a new state of the art result 58.9\% accuracy (59.2\% on the development part
of the set and 58.7\% on the test part of the set).

\begin{table} [tb]
\caption{{\it Comparison and combination of models on the Microsoft Sentence Completion Challenge.}}
\vspace{2mm}
\centerline{
\begin{tabular}{|l||c|}
\hline
Architecture               &   Accuracy [\%] \\
\hline
4-gram~\cite{MSRSCC}       &   39    \\
Average LSA similarity~\cite{MSRSCC}          &   49 \\
Log-bilinear model~\cite{Mnih2012}       &   54.8  \\
RNNLMs~\cite{thesis}       &   55.4  \\
\hline
Skip-gram                  &   48.0    \\
Skip-gram + RNNLMs         &  {\bf 58.9}  \\
\hline
\end{tabular}}
\label{SCCTab}
\end{table}

%- more than 1 example of relationship to have more stability / accuracy
%- large vectors trained on a lot of data
% for other discussion - results with context length, size of projection layer, different activation functions

\section{Examples of the Learned Relationships}

% out of list detection, enlargement of list of objects, capital cities, city to state, profession of people, ...

Table~\ref{EX1} shows words that follow various relationships. We follow the approach described above: the relationship
is defined by subtracting two word vectors, and the result is added to another word. Thus for example, {\it Paris - France + Italy = Rome}.
As it can be seen, accuracy is quite good, although there is clearly a lot of room for further improvements (note that using our accuracy metric that assumes exact match, the results in Table~\ref{EX1} would score only about 60\%).
We believe that word vectors trained on even larger data sets with larger dimensionality will perform significantly better,
and will enable the development of new innovative applications.
Another way to improve accuracy is to provide more than one example of the relationship. By using ten examples instead of one to form the relationship vector (we average the individual vectors together),
we have observed improvement of accuracy of our best models by about 10\% absolutely on the semantic-syntactic test.

It is also possible to apply the vector operations to solve different tasks. For example, we have observed
good accuracy for selecting out-of-the-list words, by computing average vector for a list of words, and finding the most distant word vector.
This is a popular type of problems in certain human intelligence tests. Clearly, there is still a lot of discoveries to be made using these techniques.

\begin{table} [tb]
\caption{ {\it Examples of the word pair relationships, using the best word vectors from Table~\ref{Acc2} (Skip-gram model trained on 783M words with 300 dimensionality).}}
\vspace{2mm}
\centerline{
\begin{tabular}{|c||c|c|c|}
\hline
Relationship            &  Example 1             &  Example 2             &   Example 3           \\
\hline
France - Paris          &   Italy: Rome & Japan: Tokyo  & Florida: Tallahassee    \\
big - bigger            & small: larger & cold: colder  & quick: quicker \\
Miami - Florida         & Baltimore: Maryland  & Dallas: Texas     & Kona: Hawaii    \\
Einstein - scientist    & Messi: midfielder & Mozart: violinist & Picasso: painter \\
Sarkozy - France        & Berlusconi: Italy & Merkel: Germany   & Koizumi: Japan  \\
copper - Cu             & zinc: Zn          & gold: Au          & uranium: plutonium    \\
Berlusconi - Silvio     & Sarkozy: Nicolas  & Putin: Medvedev   & Obama: Barack   \\
Microsoft - Windows     & Google: Android   & IBM: Linux        & Apple: iPhone   \\
Microsoft - Ballmer     & Google: Yahoo     & IBM: McNealy      & Apple: Jobs     \\
Japan - sushi           & Germany: bratwurst & France: tapas    & USA: pizza      \\
\hline
\end{tabular}}
\label{EX1}
\end{table}

% TOP 3:
%France - Paris, Italy: Rome, Amsterdam, Berlin; Japan: Tokyo, Soul, Osaka; Florida: Tallahassee, Miami, Louisiana
%big - bigger, small: larger, smaller, large; cold: colder, warmer, drier; quick: quicker, faster, sharper
%Miami - Florida, Baltimore: Maryland, Texas, Missouri; Dallas: Texas, Arkansas, Amarillo; Kona: Hawaii, Maui, Oahu
%Einstein - scientist, Messi: midfielder, playmaker, striker; Mozart: violinist, Brahms, pianist; Picasso: painter, sculptor, curator
%Sarkozy - France, Berlusconi: Italy, Spain, Belgium; Merkel: Germany, Austria, Italy; Koizumi: Japan, Italy, Germany
%copper - Cu, zinc: Zn, Pb, Au; gold: Au, Zn, gpt; uranium: plutonium, reprocessing, enrichment
%Berlusconi - Silvio, Sarkozy: Nicolas, Chirac, Fillon; Putin: Medvedev, Vladimir, Dmitry; Obama: Barack, Bush, Clinton

%Microsoft - Windows, Google: Android, Chrome, toolbar; IBM: Linux; Apple: iPhone, iPad, Android

\section{Conclusion}

% *** TODO: add stronger claims about what we can do with highly regular word vectors
% *** mention that absolute accuracy on our test set is not the crucial thing in this paper: we can get better by doing certain tricks... the point of the test set is to distinguish between really bad and good models

In this paper we studied the quality of vector representations of words derived by various models on a collection of syntactic and semantic language tasks.
We observed that it is possible to train high quality
word vectors using very simple model architectures, compared to the popular neural network models (both feedforward and recurrent). Because of the much lower computational complexity,
it is possible to compute very accurate high dimensional word vectors from a much larger data set. Using the DistBelief distributed framework, it should be possible to train
the CBOW and Skip-gram models even on corpora with one trillion words, for basically unlimited size of the vocabulary. That is several orders of magnitude larger than the best previously published results for similar models.

An interesting task where the word vectors have recently been shown to significantly outperform the previous state of the art is the SemEval-2012 Task 2~\cite{Jurgens}. The publicly available
RNN vectors were used together with other techniques to achieve over 50\% increase in Spearman's rank correlation over the previous best result~\cite{NAACL2}.
The neural network based word vectors were previously applied to many other NLP tasks, for example sentiment analysis~\cite{sentiment} and paraphrase detection~\cite{paraphrase}. It can be expected
that these applications can benefit from the model architectures described in this paper.

Our ongoing work shows that the word vectors can be successfully applied to automatic extension of facts in Knowledge Bases, and also for verification of correctness of existing facts. Results from machine
translation experiments also look very promising.
In the future, it would be also interesting to compare our techniques to
Latent Relational Analysis~\cite{Turney} and others.
We believe that our comprehensive test set will help the research community to improve the existing techniques for estimating the word vectors.
We also expect that high quality word vectors will become an important building block for future NLP applications.

\section{Follow-Up Work}

After the initial version of this paper was written, we published
single-machine multi-threaded C++ code for computing the word vectors,
using both the continuous bag-of-words and
skip-gram architectures\footnote{The code
is available at \url{https://code.google.com/p/word2vec/}}.
The training speed is significantly higher
than reported earlier in this paper, i.e. it is in the order of
billions of words per hour for typical
hyperparameter choices. We also published more than 1.4 million
vectors that represent named entities, trained on more than 100
billion words.
Some of our follow-up work will be published in an upcoming NIPS 2013
paper~\cite{NIPS2013}.

%\subsubsection*{Acknowledgments}

\bibliography{strings,refs}

\end{document}